\documentclass[conference]{IEEEtran}
\IEEEoverridecommandlockouts
\usepackage{cite}
\usepackage{amsmath,amssymb,amsfonts}
\usepackage{algorithmic}
\usepackage{graphicx}
\usepackage{textcomp}

\usepackage{graphics} 
\usepackage{epsfig} 
\usepackage{mathptmx} 
\usepackage{times} 
\usepackage{color}
\usepackage[table]{xcolor}
\usepackage{url}
\usepackage{tabularx}
\usepackage{fancyhdr}

\fancypagestyle{FirstPage}{
}

\begin{document}

\newcommand{\zhi}[1]{\textcolor{black}{#1}}
\newcommand{\iarI}[1]{\textcolor{black}{#1}}
\newcommand{\iarII}[1]{\textcolor{black}{#1}}

\title{Human-centered Benchmarking for Socially-compliant Robot Navigation
\thanks{This work was supported by the Bourgogne-Franche-Comt\'e regional research project LOST-CoRoNa.\newline
    $^1$UTBM, CIAD UMR 7533, F-90010 Belfort, France. {\tt\small firstname.lastname@utbm.fr}\newline
    $^2$Center for Bioinformatics, Saarland Informatics Campus, Saarbr\"ucken, Germany. {\tt\small lish00001@stud.uni-saarland.de}\newline
    $^*$Corresponding Author.}
}

\author{Iaroslav Okunevich$^{1}$, Vincent Hilaire$^{1}$, Stephane Galland$^{1}$,\\
Olivier Lamotte$^{1}$, Liubov Shilova$^{2}$, Yassine Ruichek$^{1}$, and Zhi Yan$^{1*}$
}

\maketitle

\begin{abstract}
Social compatibility is one of the most important parameters for service robots.
It characterizes the quality of interaction between a robot and a human. In this paper, a human-centered benchmarking framework is proposed for socially- compliant robot navigation.
In an end-to-end manner, four open-source robot navigation methods are benchmarked, two of which are socially-compliant.
All aspects of the benchmarking are clarified to ensure the reproducibility and replicability of the experiments.
The social compatibility of robot navigation methods with the Robotic Social Attributes Scale (RoSAS) is measured.
After that, the correspondence between RoSAS and the robot-centered metrics is validated.
Based on experiments, the extra robot time ratio and the extra distance ratio are the most suitable to judge social compatibility. 
\end{abstract}

\begin{IEEEkeywords}
Social navigation, human-robot interaction, benchmarking
\end{IEEEkeywords}

\section{Introduction}

\thispagestyle{FirstPage}

The development of computing and sensing technologies allows us to apply mobile robotic systems in different environments.
Robot behavior is especially important in an environment with human presence, such as in the case of mobile robots for \zhi{emerging} logistic~\cite{fukui2020hangrawler} or disinfection~\cite{perminov2021ultrabot} purposes.
In these cases, socially-compliant robot navigation~\cite{gao2022evaluation} is one of the main requirements that guarantees a high-quality human-robot interaction (HRI).

Although robot systems perform relatively well, people still tend to fear them, which negatively affects mental health and decreases the productivity of workers~\cite{sahin2022evaluation}.
The problem behind fear is the lack of understanding of robot behavior~\cite{akalin2022you}.
Robotic intelligence is different from that of humans, and human-robot interaction is limited in ways of communication compared to human-human interaction. 
People feel safer in the presence of other people, thus preferring them to robots as their working partners. The feeling of safety comes from the belief that people's behavior is more predictable. Similarly, one generally feels uneasy when communicating with a drunk person, as alcohol makes their behavior unpredictable. 

\begin{figure}[t]
  \centering
  \includegraphics[width=\columnwidth]{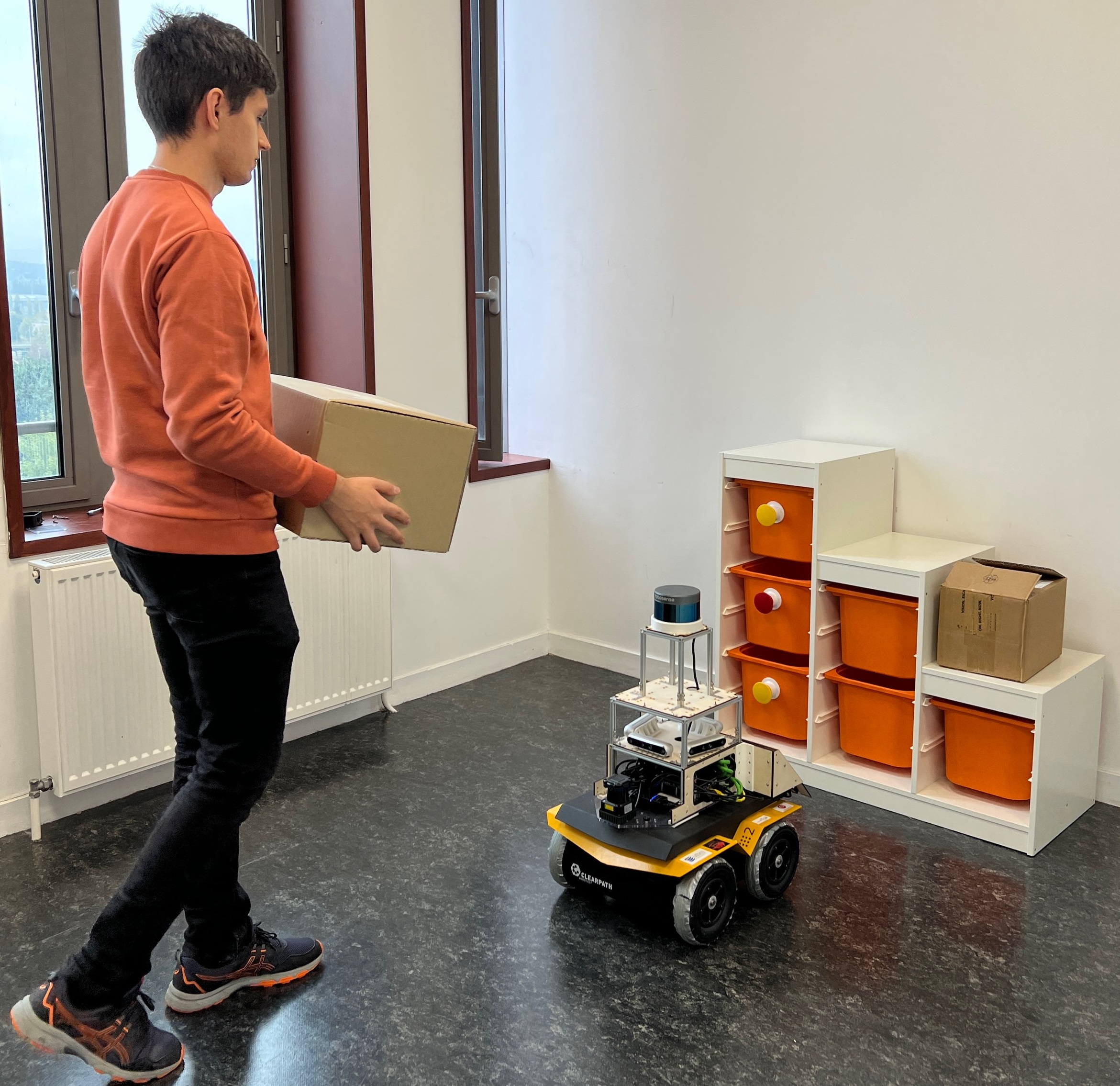}
  \caption{The experiment to examine the social compatibility of robot navigation. The person works in a room, while the mobile robot moves nearby. The logistic operations at a warehouse are an example of a real scenario, where human workers and autonomous mobile robots need to collaborate with each other and navigate in a shared space.}
  \label{fig:experiment}
\end{figure}

To make the behavior of the robot more understandable, one could apply different engineering solutions.
In addition to the sensors necessary to perceive the world,
the robot can be equipped with mechanical elements to show its behavior or intention, such as the light~\cite{fernandez2018passive} and sound~\cite{shrestha2018communicating} signaling system or an additional screen~\cite{hart2020using}.
Another way to reduce robot fear is to improve the quality of navigation algorithms.
This implies that the robot tries to follow the unspoken social rules that people have in their regular life.
For instance, the left- and right-hand rules to avoid collisions~\cite{chen2017socially}, social zones around people~\cite{truong2016dynamic}, and navigation through pedestrian flow~\cite{morales2018personal}.

However, evaluating the social effectiveness of socially-compliant navigation methods can be challenging.
Many studies~\cite{chen2017socially,truong2016dynamic} apply robot-centered metrics (RCM) to assess the quality of the social part of navigation methods.
These metrics are numerical and usually measure robot functionality as a reference. For example, the speed of the robot or the length of the traveled path.
As fear is not a numerical parameter, scientists also need to use psychological metrics to assess the acceptance of the robot by people. The latter can be regarded as compatibility from a robot perspective.
We suggest therefore to use ``social compatibility'' (SC), rather than ``social acceptance'' used in the literature, which characterizes the social effectiveness of robotic navigation methods. 
The high SC value of a navigation method implies that, in a social environment, a robot moves in an efficient, safe and socially acceptable manner~\cite{gao2022evaluation}.

One of the most popular approaches to measure SC is to invite people to participate in an experiment (such as that shown in Fig.~\ref{fig:experiment}) and then conduct a questionnaire for the participants. As questions are used as metrics to evaluate human feelings, they are called human-centered metrics (HCM).
However, in different articles various metrics and experimental settings are applied to assess the interaction between robots and humans. Consequently, reproducing these experiments is often not straightforward, making comparisons between different methods tricky.

The contributions of this paper are twofold.
  \begin{itemize}
  \item We propose an end-to-end human-centered benchmarking framework. To confirm our idea, we benchmark four open-source robot navigation methods under the proposed framework. Two of these methods have been developed to be socially-compliant.
  All experimental settings and parameters are clearly stated to ensure the reproducibility and repeatability
  of the experiments. The software-hardware integration scheme is publicly available to the community\footnote{\scriptsize{\url{https://github.com/Nedzhaken/human_aware_navigation}}}.
  \item We evaluate different methods using both HCM and RCM and report the experimental results. We gain insight that some RCMs are suitable for assessing SC while others are not, if considered for HCM. This provides a basis for clarifying the connection between RCM and HCM.
  
\end{itemize} 

\section{Related Work}

Much work has been done on socially aware robot navigation, as well as interaction between humans and autonomous mobile robots.
However, the applied experiment conditions (e.g. hardware, software, environment, etc.) and metrics to measure method performance vary from paper to paper significantly.

\cite{chen2017socially} focused on a multi-agent collision avoidance algorithm that exhibits socially-compliant behavior.
The authors trained their algorithm in a reinforcement learning framework and compared it with two algorithms in simulation.
They chose three performance metrics: 1) average extra time to reach the goal; 2) minimum separation distance to other agents; 3) relative preference between left-handedness and right-handedness.
Although the experiment in real life proved that the method developed was safe, the work did not show the opinion of the people about the behavior of the robot.
\cite{lu2013towards} compared standard and social navigation strategies for efficient robot behavior. 
For a person and a robot moving in the corridor, the following metrics were recorded:
1) the speed of the robot and the person during the experiment.
Higher speed indicates a more efficient HRI.
It was shown to be a useful metric to measure the difference in HRI representing the changing human behavior;
2) the signaling distance between the person and the robot.
For the human, it was measured when the person started to change their trajectory to react to the robot.
For the robot, it was measured when the robot started to avoid the person.
This metric was shown to be suitable for a perception system but not for HRI.

Another way to evaluate socially aware robot navigation is to use simulations. This has the advantage of repeatability of the experimental conditions for each evaluated navigation method. In addition, simulated experiments often do not require real participants, which decreases the cost of the study.~\cite{biswas2022socnavbench} presented a grounded simulation framework to evaluate social navigation. This simulator included pre-recorded pedestrian trajectory datasets in different scenarios. Despite the effectiveness of the proposed framework, the simulator included only RCM and could not provide any information on HCM.

\cite{bartneck2009measurement} developed a 29-question HRI measurement questionnaire to assess how humans feel about robots.
Questions were asked in five groups: anthropomorphism, animacy, likeability, perceived intelligence, and perceived safety. The answers are ranked from 1 to 5, with 1 being the worst and 5 being the best opinion mark.
This questionnaire has been used as a baseline for numerous questionnaires in HRI research~\cite{mavrogiannis2019effects}.
However,~\cite{carpinella2017robotic} criticized the Godspeed questionnaire~\cite{bartneck2009measurement}.
Through the exploratory factor analysis (EFA), it was shown that the Godspeed questionnaire has been loaded onto three unique factors, while originally this questionnaire was designed for the five factors/groups.
Therefore, based on the Godspeed questionnaire, Carpinella \emph{et al.} developed RoSAS.
It consisted of 18 questions, which were chosen from the psychological literature on social cognition. Despite these questionnaires being one of the ways to represent HRI, their application leads to limited autonomy, since a robot itself cannot assess it.

\cite{mavrogiannis2019effects} presented the design of the user study for the experimental evaluation of mobile robot navigation strategies in human environments.
The authors applied different RCM to define the most suitable navigation strategies for HRI, such as average acceleration and energy, minimum distance between robots and humans, irregularity of the path, efficiency of the path, time spent per unit of length of the path, and topological complexity.
After the experiment, the participants evaluated HRI during the experiment through a questionnaire.
The combination of the results of two different types of metrics allowed HRI measurement by RCM and confirmed the results by comparing the responses to the questionnaire (i.e. HCM). The work provides immensely valuable input regarding the evaluation of mobile robot navigation strategies in a controlled lab environment. However, it could also be noted that the questionnaire used was later criticized by~\cite{carpinella2017robotic}. \cite{lo2019perception} studied how different robot navigation strategies are perceived by users in terms of comfort, safety, and awareness. Their results demonstrated some correlation between safety and comfort and the distance between the robot and the pedestrian when the robot passed the intersection.

From the survey, it became clear that there is still much to be done about the benchmarking methods and standardizable metrics for socially-compliant navigation.
The existing evaluation mainly uses RCM. However, it is not completely clear how these metrics reflect the SC.
Furthermore, the lack of necessary experimental information makes benchmarking of the community difficult. The status quo drives us to develop reproducible experiments based on standardizable processes to accelerate the development and comparison of relevant methods in our community. In the current work, our aim is to develop such an experiment and explore the correlation between RCM and HCM for the SC parameter.

\section{Benchmarking Framework}

HRI benchmarking is often very challenging.
This is due to, on the one hand, the increasing complexity of the robotic system (both hardware and software) and, on the other hand, the unforeseeable and unpredictable behavior of different participants with different understandings of the experimental procedures, which makes benchmarks difficult to reproduce.
To this end, we propose an end-to-end benchmarking framework (similar to black-box testing in software engineering), focusing on human-centricity that allows rapid and efficient evaluation and comparison of the performance of different socially-compliant navigation methods, by clearly defining experimental scenarios and evaluation metrics.
Applying HCM ensures that human opinion is one of the criteria of evaluation, which makes our framework human-centered.
Moreover, we propose to divide the experiment into explicit and as small steps as possible, ideally consisting of simple motion or action primitives, which make it easier to reproduce and avoid any ambiguity.
For example, the instruction to a person could be ``go straight forward for three meters at normal speed to point B'' rather than ``go to point B''.
Based on this principle, we propose the following experimental design.

\subsection{Experiment Design}

\begin{figure}[t]
  \centering
  \includegraphics[width=\columnwidth]{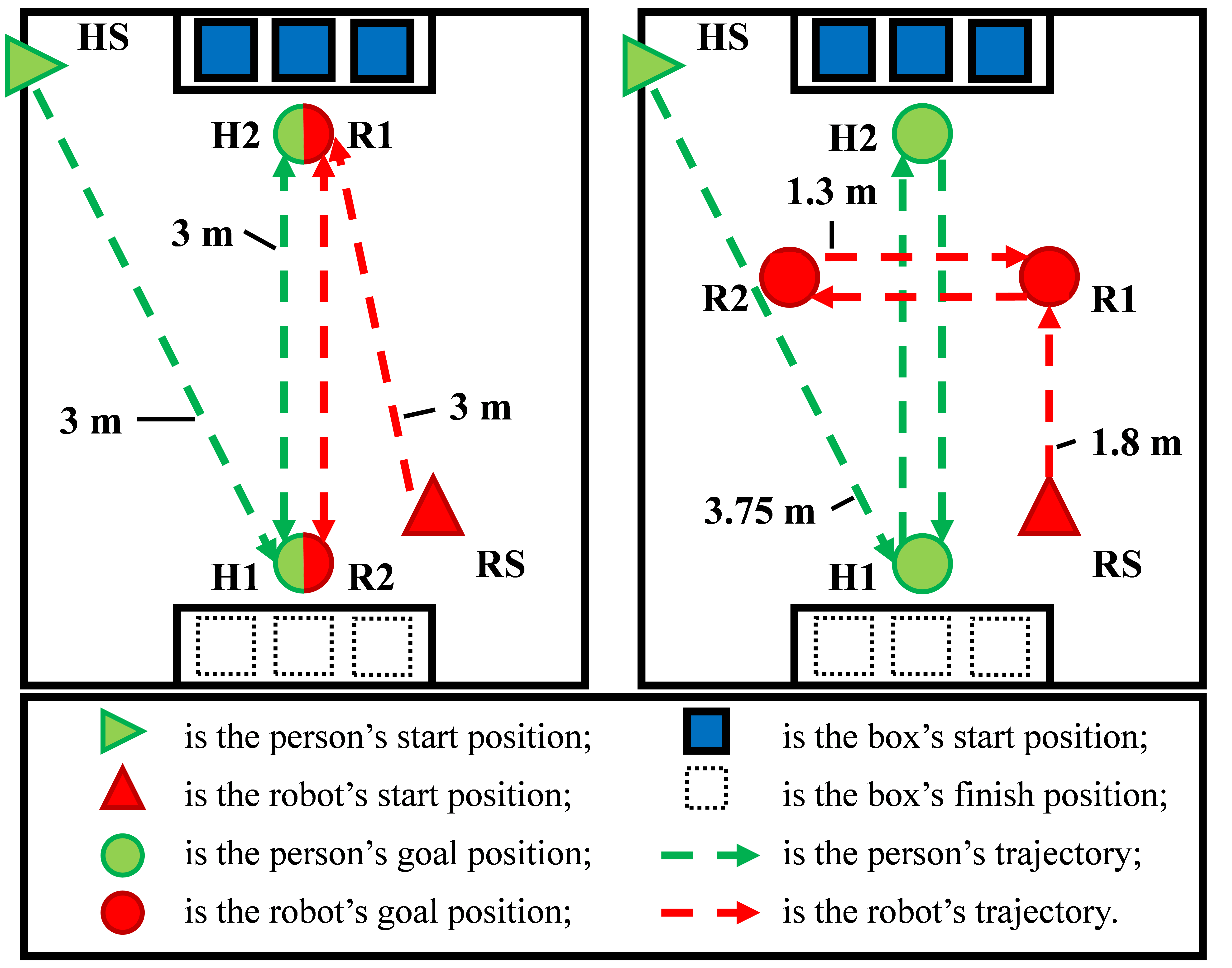}
  \caption{Our reproducible experiment design. Shown on the left is the case where the robot's trajectory is coinciding with the human's. The coordinates of positions H2-R1 and H1-R2 are equal to ensure the crossing of the robot's and human's trajectories. Shown on the right is the case where the robot moves perpendicular to the pedestrian.}
  \label{fig:scheme}
\end{figure}

Unlike the non-object experiments commonly seen in the literature~\cite{chen2017socially, mavrogiannis2022winding, pirk2022protocol, cui2021learning, repiso2020people, kretzschmar2016socially, lo2019perception}, our experiment required humans to move cartons. This task was inspired by the industrial example in which workers carry boxes in factories, warehouses, or supermarkets.
We tried to reproduce the situation in which a person should complete a working task in the presence of the robot.
This setting helps us to avoid bias in HCM results. According to research in the field of sociology, people are less likely to pay attention to robots when they concentrate on their tasks~\cite{lo2014influence}. Therefore, experiments can provide an objective and impartial assessment of SC performance, which is beneficial for comparing different methods.
Specifically, in a 2.5$\times$4~$m$ room, trial participants were asked to carry three cartons from one side of the room to the other (see Fig.~\ref{fig:scheme}).
During this period, the robot moved in the shared space.
The robot's acceleration and maximum velocity of the robot were set to 0.3~$m/{s^2}$ and 0.3~$m/s$, respectively.
Humans were told to move at normal speed.
The evaluation of the socially-compliant navigation methods included two parts: the robot path being coinciding or perpendicular to the pedestrian.
We wanted to test navigation methods in three general ways of social interaction of a mobile robot with a human: \emph{passing}, \emph{crossing}, and \emph{overtaking}~\cite{chen2017socially}. The robot movement along the human path was used to simulate the passing and overtaking scenario, and the perpendicular robot movement was used for the crossing scenario.
\zhi{The passing and crossing movements can be performed by both the robot and the human. The overtaking movement was performed only by a human, as the robot's speed was chosen to be low to decrease the influence of the velocity on SC.}
To make our experiments reproducible and to facilitate the comparison of results between different methods, we next describe the full implementation details.

The initial position of the person was at the entrance of the room, denoted as HS.
The person was asked first to reach H1 and then H2, walking in a straight line.
When reaching H2, people were asked to pick up a box and take it to H1 to drop it off.
This process was repeated until all cartons were transported to H1 and the experiment ended.
On the other hand, the starting position of the robot was in the opposite corner of the entrance to the room, marked RS.
Similarly, the robot first moved to R1 and then went back and forth between R1 and R2 four times to ensure that the person completed the task within its moving time.
As shown in Fig.~\ref{fig:scheme}, the robot moved between R1 and R2 following the same trajectory as H1-H2 or perpendicular to H1-H2.
When the robot finally reached R1, we collected the experimental RCM.

The moderately sized workspace ensures actual HRI and reliable robot navigation and allows experiments to be easily reproduced at other places. The distances between human positions were chosen to ensure the naturalness of human behavior. In the first case, the robot and human waypoints were in the same location (R1-H2 and H1-R2) to ensure that a participant interacted with the robot and did not ignore it. During the perpendicular movements of the robot, the human could solve the task without interaction with the robot. The close positioning of R1-R2 in this case increased the probability of the intersection of trajectories.

\subsection{Human-centered Metrics}

In principle, RCM alone cannot fully describe SC, as they do not reflect people's subjective feelings about the robot's behavior. To assess human opinions, we \zhi{adopted} the \zhi{aforementioned} RoSAS questionnaire. It includes 18 questions\footnote{\scriptsize\url{https://github.com/Nedzhaken/human_aware_navigation}}, each of which is answered on a scale of 1 to 9. The questions are divided into three underlying factors: warmth, competence, and discomfort. The questionnaire provides a psychometrically validated and standardized measure of HRI. The RoSAS was applied to measure social perceptions of human, robot and blended human-robot faces~\cite{carpinella2017robotic} or human-to-robot handovers~\cite{pan2018evaluating}. We \zhi{innovatively} applied the RoSAS to measure the SC of the robot navigation methods. Moreover, we wanted to demonstrate that the scale was applicable in mobile robotics to assess SC as a form of HRI.

\subsection{Robot-centered Metrics}

Five RCM metrics commonly used from the literature are selected, as well as one additional metric.

\begin{enumerate}
\item \emph{The robot extra time ratio} evaluates how efficiently a robot can complete a task in an environment shared with humans~\cite{chen2017socially, everett2018motion, liu2020robot}, and is defined as:
\begin{equation}
  R^r_{extra} = T^r / T^r_h,
\end{equation}
where $T^r$ and $T^r_h$ are the time it takes the robot to complete the task without and in the presence of humans, respectively.

\item \emph{The human extra time ratio} is a human analog of the previous one. It is first proposed in this paper to assess changes in human performance when working with robots. 
It could improve our understanding of the connection between human performance and SC. 
It is defined as:
\begin{equation}
  R^h_{extra} = T^h / T^h_r,
\end{equation}
where $T^h$ and $T^h_r$ are the time it takes a human to complete the task without and in the presence of robots, respectively.

\item \emph{The extra distance ratio} evaluates system performance in terms of the distance a robot would have to travel additionally when a human is present~\cite{gao2021evaluation, mavrogiannis2019effects}, and is defined as:
\begin{equation}
  R_{dist} = D^r / D^r_h,
\end{equation}
where $D^r$ and $D^r_h$ represent the distance that the robot travels to complete a task without and in the presence of a human, respectively.

\item \emph{The success ratio} assesses the ability of a robot to complete a task without colliding with a human~\cite{chen2017socially, everett2018motion, liu2020robot}, and is defined as:
\begin{equation}
  R_{succ} = N_{succ} / N,
\end{equation}
where $N_{succ}$ represents the number of successful trials during which the robot does not hit a human and $N$ indicates the total number of trials.

\item \emph{The hazard ratio} assesses the time that a robot gets too close to a human~\cite{liu2020robot}, which is defined as:

\begin{equation}
  R_{haza} = \frac {{1}}{n} \cdot \sum_{i=1}^{n} \frac{{T^{hazard}_i}}{T^{social}_i},
\end{equation}
where $n$ is the number of people, $T^{hazard}_i$ is the duration of time when the distance between the robot and the i-$th$ person is less than the safe distance (denoted as $D_{safe}$), and $T^{social}_i$ is the duration of time when the distance between the robot and the i-$th$ person is less than the social distance (denoted as $D_{social}$).
In our experiments, $D_{safe} = 0.2~m$ and $D_{social} = 0.4~m$.

\item \emph{The deceleration ratio} evaluates a robot's ability to slow down when approaching a human~\cite{lu2013towards}, which is defined as:
\begin{equation}
  R_{dec} = \frac{{1}}{n} \cdot \sum_{i=1}^{n} \frac{{V_i}}{V^{max}},
\end{equation}
where $n$ represents the number of speed measurements when the robot is less than $D_{social}$ from the human.
$V_i$ represents the instantaneous speed of the robot at i-$th$ measurement, and $V^{max}$ is the maximum speed of the robot (0.3~$m/s$).
The maximal velocity was kept the same for all methods.
Although the different methods can work with different maximal velocities, variations in this parameter would complicate the analysis. It would be difficult to understand whether the maximal velocity or the algorithm itself affects the SC.
\end{enumerate}

\section{Experiments}

Our experiments aimed to benchmark four open-source robot navigation methods. Two of them were developed as socially-compliant. This, on the one hand, showed the effectiveness of the proposed benchmarking framework and, on the other hand, revealed the connection between RCM and HCM.

\subsection{Experimental Platform}

For the experiment, we used a mobile robotic platform. The robot chassis is a Clearpath Jackal UGV. The perception system includes four RGB-D cameras and a 3D lidar. The RGB-D cameras are placed toward all sides of the robot for a panoramic view. The 3D lidar allows people detection and tracking under different lighting conditions. The robot is equipped with a 2D lidar that has higher measurement frequency, accuracy, and resolution compared to the 3D lidar. It is beneficial for robot localization and collision-free navigation. The software system has been fully implemented in ROS~\cite{gatesichapakorn2019ros} with high modularity and is publicly available to the community.

\subsection{Evaluated Methods}

\zhi{We deployed several open-source methods and reported results on four of them.} The choice was based on two factors: 1) the method must be deployable on real robots, and 2) the effectiveness of the method must have been confirmed in its corresponding paper.
\zhi{Two of these methods are socially-compliant robot navigation methods and two are traditional navigation approaches that include only collision avoidance mechanisms.}
\begin{itemize}
\item 
  \emph{Social Navigation Layers (SNL)}\footnote{\scriptsize\url{https://github.com/DLu/navigation_layers}}~\cite{lu2014layered}: This method implements a Gaussian mixture model around the detected person on the navigation cost map.
  The extra cost area around the person makes the robot consider avoiding it when planning its path. This allows the robot to demonstrate better social attributes during navigation. Also, if the person moves, the social area grows in the direction of the movement (i.e., from a circle to an ellipse). In our experiments,
  \zhi{according to the characteristics of the working environment,}
  the social radius was set to be 0.4~$m$ centered on the person.
\item \emph{Time Dependent Planning (TDP)}\footnote{\scriptsize\url{https://github.com/marinaKollmitz/human_aware_navigation}}~\cite{kollmitz2015time}: This method is similar to SNL, except that the social area is no longer limited to a person's current location, but also includes their predicted location several time steps in the future, based on a constant velocity model.
\item \emph{Collision Avoidance with Deep Reinforcement Learning (CADRL)}\footnote{\scriptsize\url{https://github.com/mit-acl/cadrl_ros}}: This method is the underlying implementation of the well-known SA-CADRL (socially aware CADRL)~\cite{everett2018motion}, while the latter has not been ROSified. However, it is still considered a baseline, as collision avoidance is one of the most fundamental elements in the social properties of robot navigation.
\item \emph{move\_base (MB)}\footnote{\scriptsize\url{https://github.com/ros-planning/navigation}}: This is a basic component provided by the ROS navigation stack and does not contain any socially-compliant modules.
\end{itemize}

Additionally, we added the human-human interaction to understand the difference between a robot and a human interaction in the terms of HCM.

\begin{itemize}
\item \emph{Human-human interaction (HH)}: In this case, the robot is replaced by a human who performs the task assigned to the robot, that is, moving from one point to another.
\end{itemize}

The results of the RCM were recorded during the execution of the above methods by the robot, and the participants were asked to complete the questionnaire after each method to assess the HCM.

\subsection{Participants}

The recruitment was carried out within the University of Technology of Belfort-Montbéliard (UTBM) in France. Twenty volunteers (14 men, 6 women), aged 18 to 39 years [\emph{M} = 27.10, \emph{SD} = 5.30] participated in the experiment. Participants were not rewarded in this research.
To avoid carry-over effects, the methods of the experiment were counterbalanced among participants by applying a Latin square design~\cite{pan2018evaluating}.

\subsection{Experimental Results}

As RoSAS had not been used before to measure SC of a mobile robot, we performed an internal consistency (IC) test, which allows us to confirm the results of the EFA performed in the original investigation~\cite{carpinella2017robotic}.
Specifically, the IC measures how closely the RoSAS questions match three factors (warmth, competence, and discomfort) by applying the data from our experiment. For the test, Cronbach's alpha should be more than $0.90$ to represent high IC~\cite{nunnally1978psychometric}. Cronbach's alphas of warmth ($\alpha_{Cronbach} = 0.94$), competence ($\alpha_{Cronbach} = 0.94$), and discomfort ($\alpha_{Cronbach} = 0.92$) satisfied this condition. Thus, the factors have relatively high IC with their respective questions. For the analysis of RoSAS, six questions were averaged that comprise the dimensions of warmth, competence, and discomfort. The warmth factor includes the items: happy, feeling, social, organic, compassionate, and emotional. The competence factor includes the following elements: capable, responsive, interactive, reliable, competent, and knowledgeable. The discomfort factor includes items: scary, strange, awkward, dangerous, awful, and aggressive.
The one-way ANOVA results (see Table~\ref{tabl:T1}) show that there is a statistically significant difference between the methods evaluated for each HCM and applied RCM ($p<0.05$) except for $R^h_{extra}$.

\begin{table}[ht]
  \centering
  \caption{ANOVA results of applied HCM and RCM}
  \label{tabl:T1}
  \begin{tabular}{|l||c|c|c|}
    \hline
    \textbf{Metric} & \textbf{\shortstack{\\Sum Sq}} & \textbf{\shortstack{\\F value}} & \textbf{\shortstack{\\p}}\\
    \hline
    \hline
    Warmth &\shortstack{\\250.134} & \shortstack{\\28.203} & \shortstack{\\$<$0.001}\\
    \hline
    Competence &\shortstack{\\173.484} & \shortstack{\\20.495} & \shortstack{\\$<$0.001}\\
    \hline
    Discomfort &\shortstack{\\110.194} & \shortstack{\\10.317} & \shortstack{\\$<$0.001}\\
    \hline
    $R^r_{extra}$ &\shortstack{\\0.957} & \shortstack{\\21.608} & \shortstack{\\$<$0.001}\\
    \hline
    $R^h_{extra}$ &\shortstack{\\0.045} & \shortstack{\\1.501} & \shortstack{\\0.22}\\
    \hline
    $R_{dist}$ &\shortstack{\\0.041} & \shortstack{\\3.025} & \shortstack{\\0.035}\\
    \hline
    $R_{succ}$ &\shortstack{\\0.459} & \shortstack{\\3.435} & \shortstack{\\0.021}\\
    \hline
    $R_{haza}$ &\shortstack{\\0.104} & \shortstack{\\4.052} & \shortstack{\\0.010}\\
    \hline
    $R_{dec}$ &\shortstack{\\2.626} & \shortstack{\\166.332} & \shortstack{\\$<$0.001}\\
    \hline
  \end{tabular}
\end{table}

\begin{figure}[t]
  \centering
  \includegraphics[width=\columnwidth]{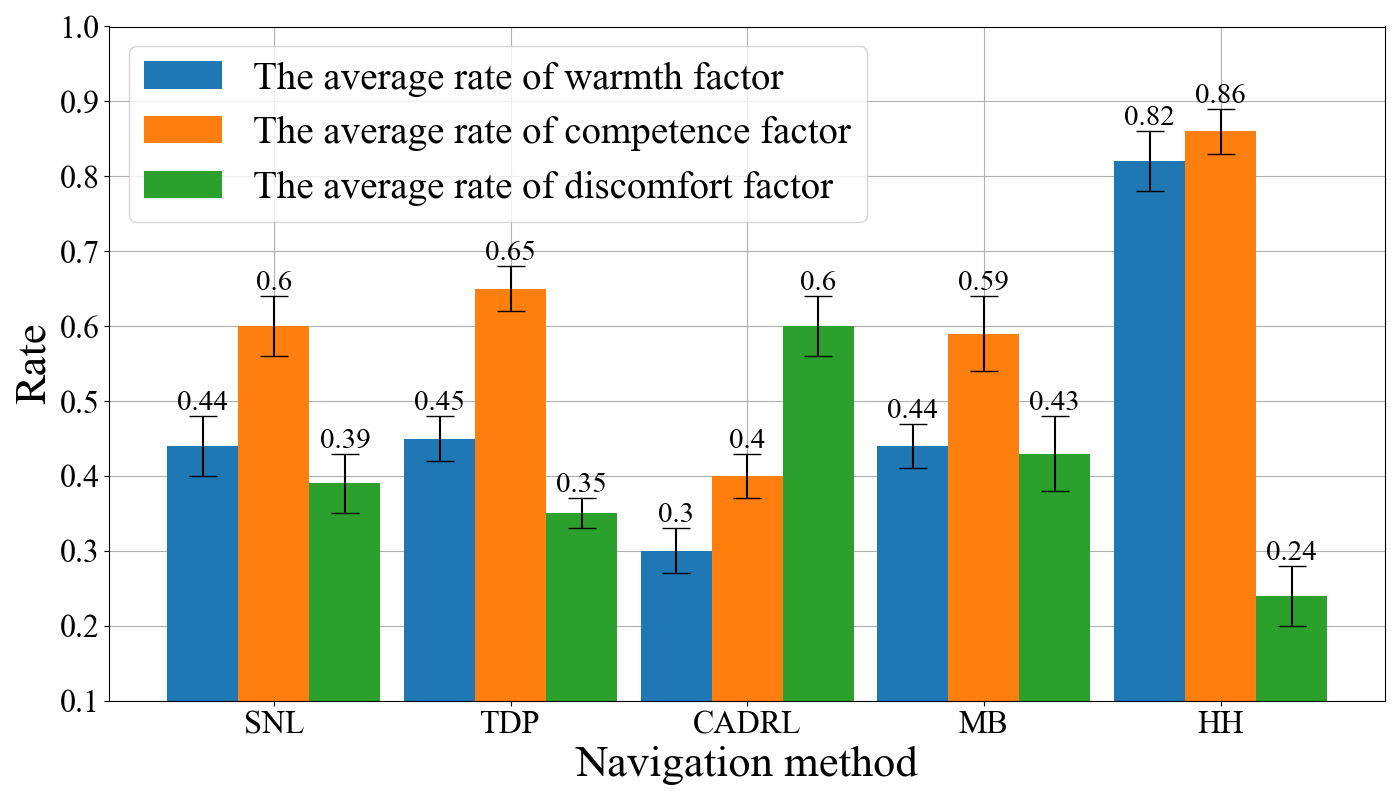}
  \caption{Experimental results of RoSAS. The bars represent the average values of 3 questionnaire factors (warmth, competence, discomfort), normalized to $[0, 1]$ with standard error ($N = 20$). 
  }
  \label{fig:metrics}
\end{figure}

Fig.~\ref{fig:metrics} summarizes the normalized HCM results. The blue, orange, and green bars represent \zhi{respectively} the average rates of the warmth, competence, and discomfort factor of RoSAS.
\zhi{It can be seen that TDP performs best in experiments involving the robot.}
This is reasonable, as this method is the only one with pedestrian prediction capability, which also confirms the importance of robot foresight in socially-compliant navigation.
The reason for the worst CADRL performance is the freezing movement of the robot during the experiment. The reason for that is the implementation of \zhi{the open-source version of the algorithm.}
Therefore, it leads to \zhi{aggressive motion and freezing of the robot,} therefore to low rates of warmth and competence and a high rate of discomfort.
\zhi{Instead,} HH scores for warmth and competence are much higher and for discomfort much lower than in other robot-\zhi{involved} methods.
This reflects the general understanding that people still find other people more socially acceptable than robots. 
\iarII{The difference between SNL and MB is only the implementation of social zones in SNL. The close values of the warmth and competence factors of MB and SNL demonstrate that these social zones influence exclusively the discomfort factor.}

The results of the experiment are presented in Table~\ref{tabl:T3}. The gray row shows that $R^h_{extra}$ provides values that do not vary significantly among the methods. Red and green cells are \zhi{respectively} the worst and best results of a metric in terms of SC.
In terms of HCM, the CADRL with its freezing movements can be seen as the worst and the TDP with pedestrian prediction capability as the best method. RCM partially follows this trend.
On the one hand, three out of five RCMs were indeed the worst for CADRL. On the other hand, $R^r_{extra}$ and $R_{succ}$ demonstrated the method to be the best. This means that, while the robot did not pose a real danger to people and did not spend extra time with them, it was still perceived as the most uncomfortable to work with.
For TDP, only $R_{dist}$ reached the best value. In line with the HCM results, this metric has the highest value in TDP and the lowest value in CADRL.

\begin{table}[b]
\setlength{\arrayrulewidth}{0.2mm}
  \centering
  \caption{Experimental results of RCM and HCM}
  \label{tabl:T3}
  \begin{tabular}{|l||c|c|c|c|}
    \hline
    \textbf{Metric} & \textbf{\shortstack{\\SNL}} & \textbf{\shortstack{\\TDP}} & \textbf{\shortstack{\\CADRL}} & \textbf{\shortstack{\\MB}}\\
    \hline
    \hline
    Warmth &\shortstack{\\0.44} & \cellcolor{green!25}\shortstack{\\0.45} & \cellcolor{red!25}\shortstack{\\0.30}&\shortstack{\\0.44}\\
    \hline
    Competence &\shortstack{\\0.60} & \cellcolor{green!25}\shortstack{\\0.65} & \cellcolor{red!25}\shortstack{\\0.40}&\shortstack{\\0.59}\\
    \hline
    Discomfort &\shortstack{\\0.39} & \cellcolor{green!25}\shortstack{\\0.35} & \cellcolor{red!25}\shortstack{\\0.60}&\shortstack{\\0.43}\\
    \hline       
    $R_{haza}$ &\shortstack{\\0.59} & \shortstack{\\0.57} &\cellcolor{red!25} \shortstack{\\0.65}&\cellcolor{green!25}\shortstack{\\0.56}\\
    \hline 
    \rowcolor{gray!25}$R^h_{extra}$ &\shortstack{\\0.9} & \shortstack{\\0.88} & \shortstack{\\0.94}&\shortstack{\\0.87}\\
    \hline       
    $R_{dist}$ &\shortstack{\\0.96} & \cellcolor{green!25}\shortstack{\\1.00} & \cellcolor{red!25}\shortstack{\\0.95}&\shortstack{\\0.97}\\
    \hline
    $R_{dec}$ &\shortstack{\\0.56} & \shortstack{\\0.58} & \cellcolor{red!25}\shortstack{\\0.17}&\cellcolor{green!25}\shortstack{\\0.61}\\
    \hline
    $R^r_{extra}$ &\shortstack{\\0.77} & \cellcolor{red!25}\shortstack{\\0.74} & \cellcolor{green!25}\shortstack{\\1.00}&\shortstack{\\0.83}\\
    \hline  
    $R_{succ}$ &\shortstack{\\0.92} & \shortstack{\\0.85} & \cellcolor{green!25}\shortstack{\\1.00}&\cellcolor{red!25}\shortstack{\\0.8}\\
    \hline
  \end{tabular}
\end{table}

$R^r_{extra}$ shows the inverse relation to HCM, which allows the application of the inverse value of $R^r_{extra}$ to measure SC. The reason for the lowest value $R^r_{extra}$ of TDP is the pause during movements. The mobile robot with the pedestrian prediction capability prefers to wait while the person liberates the path of the robot than trying to avoid them. In this case, the robot spends more time finishing the task but crosses fewer distances and seems to be better accepted by people. As $R^r_{extra}$ and $R_{dist}$ have low correlation coefficients (see Table~\ref{tabl:T2}), the relationship between these RCM and HCM is likely non-linear.

\begin{table}[t]
  \centering
  \setlength{\arrayrulewidth}{0.2mm}
  \caption{Correlation coefficients of RCM to HCM}
  \label{tabl:T2}
  \begin{tabularx}{\linewidth}{|X||c|c|c|c|c|c|}
    \hline
    \textbf{Metric} & \textbf{\shortstack{\\$R_{haza}$}} & \cellcolor{gray!25}\textbf{\shortstack{\\$R^h_{extra}$}} & \textbf{\shortstack{\\$R_{dist}$}} & \textbf{\shortstack{\\$R_{dec}$}} & \textbf{\shortstack{\\$R^r_{extra}$}} & \textbf{\shortstack{\\$R_{succ}$}} \\
    \hline
    \hline
    Warmth &\shortstack{\\-0.700} & \cellcolor{gray!25}\shortstack{\\0.148} & \shortstack{\\0.152} & \shortstack{\\-0.402} & \shortstack{\\0.114} & \shortstack{\\0.086}\\
    \hline
    Compet. &\shortstack{\\-0.620} & \cellcolor{gray!25}\shortstack{\\0.304} & \shortstack{\\0.156} & \shortstack{\\-0.195} & \shortstack{\\0.029} & \shortstack{\\0.085}\\
    \hline
    Discom. &\shortstack{\\0.454} & \cellcolor{gray!25}\shortstack{\\-0.456} & \shortstack{\\-0.197} & \shortstack{\\0.059} & \shortstack{\\-0.079} & \shortstack{\\0.022}\\
    \hline
  \end{tabularx}
\end{table}

The $R_{haza}$, $R_{dec}$, and $R_{succ}$ do not seem to match the HCM trends when comparing the methods, although the correlation coefficients for some of them are considerable. 
The values of $R_{haza}$ are similar for SNL, TDP and MB. This matches the warmth factor of HCM. As expected, the more often the robot is located near the human, the lower is the warmth and competence factors, and the greater is the discomfort. $R_{dec}$ has a trend similar to $R_{haza}$. The larger decrease in speed in close proximity to the person corresponds to a worse HCM. However, as with $R_{haza}$, the highest speed of the robot near the participants does not correspond to the best HCM.
Interestingly, $R_{succ}$ does not reflect HCM. The reason might be the low speed of the robot in the experiment, which made collisions negligible to the participants.

\zhi{Therefore, the following conclusions can be made:}
\begin{itemize}
\item TDP has the best HCM among the robot navigation methods, because of \zhi{its} pedestrian prediction capability.
\item HH interaction has higher values of HCM and therefore higher SC.
\item When people worked with the robot, they needed more time to complete the tasks (\zhi{i.e.} $R^h_{extra}<1.00$ for each method).
\item While $R^r_{extra}$ and $R_{dist}$ reflect the HCM and can be used to judge SC, other RCM do not give a clear picture of SC.
Therefore, in the experiments with mobile robots, especially when assessing human opinion is not possible, it is highly advisable to record $R^r_{extra}$ and $R_{dist}$ to judge the SC of the navigation method.
\end{itemize}

\section{Conclusions}

In this paper, we proposed a human-centered benchmarking framework for socially-compliant robot navigation with RoSAS and benchmarked four open-source approaches.
The benchmarking framework is end-to-end and explicitly provides all parameters required for the reproduction of experimental results.
This benchmark aims to evaluate the social part of a navigation method.
Only the full survey of participants, preferably conducted with a standard questionnaire such as RoSAS, can provide the full picture of SC. However, in situations where it is not possible, one could record RCM like $R^r_{extra}$ and $R_{dist}$ that reflect the SC of navigation.
We suggest to apply these two metrics for the comparison of state-of-the-art and new socially-complaint robot navigation methods in simulators.

Our future work will explore new approaches for socially-compliant navigation and continue to evaluate them under the proposed benchmarking framework.
Furthermore, our objective is to develop dependence functions for socially-compliant navigation methods from the most relevant RCM.
This task can be done using neural networks, but more training data needs to be collected.

\section*{Acknowledgment}

We thank RoboSense for sponsoring the 3D lidar, UTBM CRUNCH Lab for support in robotic instrumentation, and all those involved in the experiments.

\bibliographystyle{IEEEtran} 
\bibliography{references}

\end{document}